\documentclass{ecai}  

\usepackage{graphicx}
\usepackage{latexsym}
\usepackage{amsmath} 
\usepackage{tabularx} 
\usepackage{booktabs} 
\usepackage{moresize}
\usepackage{float}
\usepackage[numbers]{natbib}

\begin{document}

\begin{frontmatter}

\title{SalFAU-Net: Saliency Fusion Attention U-Net for Salient
Object Detection}

\author[A]{\fnms{Kassaw}~\snm{Abraham Mulat} \thanks{Corresponding Author. Email: abraham88@stu.cwnu.edu.cn}}
\author[B]{\fnms{Zhengyong}~\snm{Feng}}
\author[C]{\fnms{Tegegne}~\snm{Solomon Eshetie}}
\author[D]{\fnms{Ahmed}~\snm{Endris Hasen}}

\address[A, C]{School of Computer Science, China West Normal University, Nanchong, 637002, Sichuan, China}
\address[B]{School of Electronic and Information Engineering, China West Normal University, Nanchong, 637002, Sichuan, China;}
\address[D]{Paul C. Lauterbur Research Center, Shenzhen Institute of Advanced Technology, Chinese Academy of Science, Shenzhen, 518055, Guangdong, China}

\begin{abstract}
\normalsize Salient object detection (SOD) remains an important task in computer vision, with applications ranging from image segmentation to autonomous driving. Fully convolutional network (FCN)-based methods have made remarkable progress in visual saliency detection over the last few decades. However, these methods have limitations in accurately detecting salient objects, particularly in challenging scenes with multiple objects, small objects, or objects with low resolutions. To address this issue, we proposed a Saliency Fusion Attention U-Net (SalFAU-Net) model, which incorporates a saliency fusion module into each decoder block of the attention U-net model to generate saliency probability maps from each decoder block. SalFAU-Net employs an attention mechanism to selectively focus on the most informative regions of an image and suppress non-salient regions. We train SalFAU-Net on the DUTS dataset using a binary cross-entropy loss function. We conducted experiments on six popular SOD evaluation datasets to evaluate the effectiveness of the proposed method. The experimental results demonstrate that our method, SalFAU-Net, achieves competitive performance compared to other methods in terms of mean absolute error (MAE), F-measure, s-measure, and e-measure.
\end{abstract}

\end{frontmatter}

\section{Introduction}
\normalsize Salient object detection (SOD), also referred to as visual saliency detection is detecting the most noticeable, unique, and visually distinct objects or regions in a scene that attract the human eye \cite{borji2014salient}. The human visual perception system has an exceptional ability to rapidly recognize and focus its attention toward visually unique and prominent objects or regions within scenes \cite{ullah2020brief}. This innate capability has captivated the interest of many researchers in the field of computer vision, where the aim is to simulate this process based on the psychological and biological properties of the human visual attention system.
The goal is to identify prominent objects in images and videos that hold significant importance and valuable information.\

Given the diverse applications of SOD in various domains of computer vision, It plays a crucial role as a preprocessing step in tasks like image segmentation \cite{achanta2008salient}, \cite{qin2014integration}, \cite{fu2017object}, \cite{donoser2009saliency}, object detection \cite{borji2011scene}, \cite{rutishauser2004bottom}, \cite{ren2013region}, image captioning \cite{xu2015show}, autonomous driving \cite{pal2020looking}, and augmented reality \cite{duan2022saliency}, numerous visual saliency detection methods have been proposed. These methods aim to distinguish the most unique foreground images from less significant backgrounds. While traditional saliency detection approaches rely on low-level heuristic visual features, these methods often fail to detect salient objects in challenging scene. Recently, deep learning methods, particularly convolutional neural networks (CNNs), have exhibited exceptional efficacy across diverse computer vision tasks, including saliency detection. In contrast to traditional methods, CNN-based methods have made remarkable advancements by harnessing advanced semantic features \cite{ji2021cnn}. 
 
Due to the significant impact of representative features on algorithms performance, it is beneficial to investigate models that leverage multilevel features and contextual information to enhance saliency detection. Furthermore, despite the introduction of end-to-end models based on FCNs, there remains significance in incorporating and advancing conventional FCN models like U-Net \cite{ronneberger2015u} and its variate for the task of saliency detection.
One of the variants of U-Net that is well known for its efficacy in medical image segmentation is the Attention U-Net network \cite{oktay2018attention}, which selectively focuses on relevant regions of the input image by integrating attention mechanisms into its architecture, which improves the model's ability to capture intricate patterns and important features. The attention mechanism facilitates improved performance in tasks such as image segmentation. Drawing on its success in medical image segmentation, this study explores the application of Attention U-Net for saliency detection tasks. We added a saliency fusion module (SFM) to each decoder block of the network. This module allows us to generate saliency maps effectively, which we then concatenate each decoder's side output saliency map to get the final saliency map. The attention gate module in the proposed method helps the model learn to focus on salient features with varying sizes and shapes. In this way, SalFAU-Net has the ability to suppress irrelevant regions from an input image while emphasizing the features that are most important for saliency detection.
To summarize, the main contributions of this paper are as follows: \\
\qquad \textbf{(1)} We proposed a Saliency Fusion Attention U-Net (SalFAU-Net) for the task of visual saliency detection.\\
\hspace{1cm} \textbf{(2)} Saliency Fusion Module is added to each decoder block of the network to generate a saliency map from each decoder, and these saliency maps are concatenated together to obtain the ultimate visual representation that highlights the most important areas or objects in an image.\\
\hspace{1cm} \textbf{(3)} We conducted experiments on six publicly available challenging SOD datasets, and the results demonstrate the effectiveness of SalFAU-Net for the task of visual saliency detection.
\section{Related Works}\label{sec:related}
\normalsize Generally, saliency detection methods can be classified into two categories. These are traditional methods and deep learning-based methods. Traditional methods are based on low-level heuristic visual features such as contrast, location, and texture. Most of these methods are unsupervised or semi-supervised. Examples of traditional saliency detection methods include those based on local-contrast \cite{li2013saliency}, global-contrast \cite{yan2013hierarchical}, backgroundness-prior \cite{yang2013saliency}, center-prior \cite{xie2012bayesian}, objectness-prior \cite{li2015inner}, and others. These methods achieved good results in uncomplicated images or scenarios featuring solitary objects. Nonetheless, these methods failed to detect salient objects that are in complex scenes, low resolutions, or scenes with multiple salient objects. This limitation arises from their reliance on low-level features, which prove inadequate for addressing the complexities introduced by such challenging visual contexts. \

Recently, deep learning-based methods, particularly CNNs, have demonstrated remarkable performance across diverse computer vision tasks, including image classification \cite{lei2019dilated}, semantic image segmentation \cite{liu2021review}, and object detection \cite{xie2021oriented}.  CNNs have the capability to learn rich and hierarchical representations of input data by extracting high-level semantic features. However, in SOD, both low-level and high-level features are important for developing good visual saliency detection models. The introduction of FCNs \cite{long2015fully} has revolutionized the approach to end-to-end pixel-level saliency detection. Initially designed for semantic segmentation, FCN seamlessly combines the tasks of feature extraction and pixel label prediction in a single network structure composed of down-sampling and upsampling paths. Subsequently, numerous FCN-based visual saliency detection models have been proposed, including deep contrast learning (DCL) \cite {li2016deep}, aggregating multi-level convolutional feature framework (Amulet) \cite{zhang2017amulet}, recurrent fully convolutional networks (RFCN) \cite{wang2016saliency},  and deep uncertain convolutional features (UCF) \cite{zhang2017learning}. These advancements have notably enhanced the effectiveness of algorithms designed for visual saliency detection. Nonetheless, exploring effective FCN-based models designed for different purposes is still beneficial. U-Net is one of the most widely used networks in medical image segmentation \cite{ronneberger2015u}. following the success of U-net, numerous network variations have been introduced for different tasks. One exemplary variant of U-Net is the Attention U-Net model, which is designed for pancreas image segmentation, has shown impressive results in other tissue and organ segmentation, benefiting from the attention gate module to focus on relevant and variable size regions in an image. Most FCN-based saliency models are based on plain U-Net and have achieved remarkable performance for saliency detection. In \cite{qin2020u2}, Qin \textit{et al}. proposed a two-level nested U-structure by using a Residual U-block (RSU) as a backbone for visual saliency detection. Compared to many other networks that use pre-trained networks as backbones, U-2-Net's RSU block increases architecture depth without significantly increasing computational costs while achieving competitive performance. In \cite{han2019convolutional}, Han \textit{et al}. proposed a modified U-Net network for saliency detection, utilizing an edged convolution constraint. This variant effectively integrates features from multiple layers, reducing information loss and enabling pixel-wise saliency map prediction rather than patch-level prediction, which is common in CNN-based models. 

Although these methods based on plain U-Net achieved remarkable performance for saliency detection, their performance can be boosted by incorporating different techniques into the encoder and decoder blocks of their architecture. Recently, attention mechanisms have shown remarkable results across various computer vision applications, encompassing saliency detection. In \cite{li2020stacked}, Li \textit{et al.} proposed a U-shape network with stacked layers incorporating channel-wise attention to extract the most important channel features and effectively utilize these features by integrating a parallel dilated convolution module (PDC) and a multi-level attention cascaded feedback (MACF) module.  

In order to recurrently translate and aggregate the context features separately with various attenuation factors,  Hu \textit{et al.} \cite{hu2020sac} proposed a spatial attenuation context module. After that, the module carefully learned the weights to adaptively incorporate the collective contextual features.
In \cite{zhang2020attention}, Zhang \textit{et al}. proposed a novel approach to visual saliency detection that leverages attention mechanisms for refining saliency maps, incorporating bi-directional refinement for enhanced accuracy. The introduction of bi-directional refinement highlights the focus on comprehensive feature extraction and optimization. In \cite{zhao2019pyramid}, Zhao and Wu applied spatial attention (SA) and channel-wise attention (CA) to distinct aspects of the model. Specifically, SA was employed for low-level feature maps, while CA was incorporated into context-aware pyramid feature maps. This strategic approach aims to direct the network's focus towards the most relevant features for the given sample. 
In \cite{gong2022improved}, \textit {Gong et al} proposed an enhanced U-Net model incorporating pyramid feature attention, channel attention, and a pyramid feature extraction module to improve the performance of the U-Net backbone network.

In this research, we attempt to explore the applications of attention U-Net architecture in the realm of visual saliency detection. We added a Saliency Fusion Module (SFM) to each decoder of the network and concatenated their output to obtain the final saliency map. The attention gate module in the proposed method helps the model learn to focus on salient features with varying sizes and shapes. Thus, SalFAU-Net adeptly learns the capability to suppress irrelevant or undesirable regions within an input image while highlighting the most crucial and salient features essential for the task of saliency detection.

\section{Methodology}\label{sec:method}
\normalsize In this section, we provide a detailed description of the architecture of our proposed method. This is followed by the network supervision, the datasets and evaluation metrics used, and the implementation details.

\subsection{Architecture of SalFAU-Net} \label{salfaunet}
\begin{figure*}[h]
    \centering
    \includegraphics[width=\textwidth]{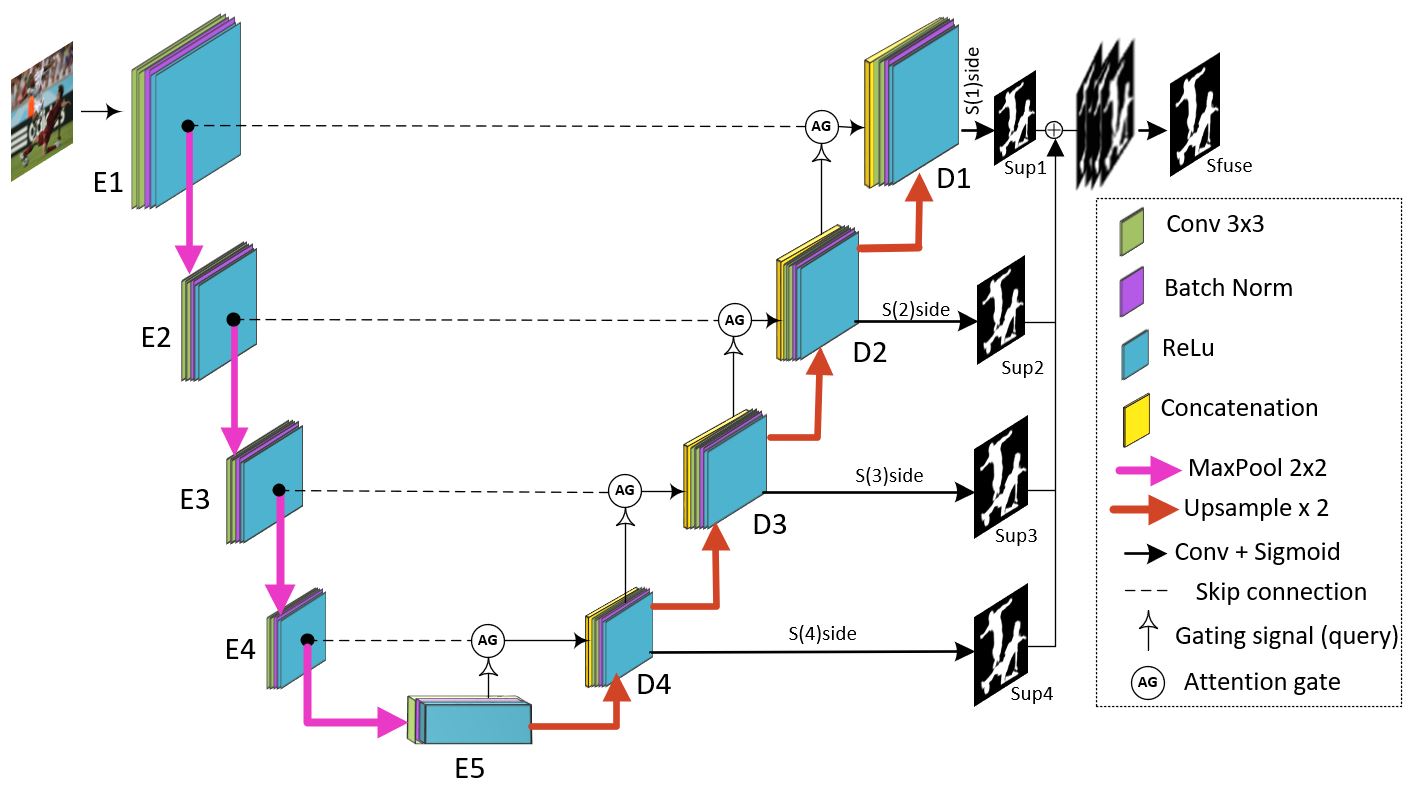}
    \caption{ Architecture of our proposed Saliency Fusion Attention U-Net (SalFAU-Net model).}
    \label{fig:model_arc}
\end{figure*}

\normalsize The proposed Saliency Fusion Attention U-Net (SalFAU-Net) for visual saliency detection in this paper mainly consists of four parts: 1) a five-level encoder block, 2) a four-level decoder block,
3) an attention gate module, and 4) a saliency fusion module. Figure \ref{fig:model_arc} shows the architecture of the proposed SalFAU-Net model. Compared with the Attention U-Net model proposed for pancreas image segmentation \cite{oktay2018attention}, we add a saliency fusion module to each decoder of the architecture and finally concatenate them together to obtain the final saliency map. \\
\subsubsection{\textbf{Encoder Block}} \label{encoder} 
\normalsize Each encoder block consists of two convolutional layers, each followed by a batch normalization layer and ReLu activation function, which increases the number of feature maps from 3 to 1024. Max pooling with a stride of 2x2 is applied at the end of every block except the last block for downsampling, reducing the image size from 288x288 to 18x18.  The encoder block progressively reduces the spatial resolution of feature maps while increasing the number of channels, capturing features at different scales. \\
\subsubsection{\textbf{Decoder Block}} \label{decoder}
\normalsize The decoder block is responsible for upsampling and generating saliency maps. It consists of an up-sampling layer followed by two convolutional layers, batch normalization, and ReLU activation function. The decoder block is connected to the attention gate block through skip connections. Each decoder block reduces the number of feature maps by two while increasing the size of the spatial resolutions from 18x18 to 288x288. The goal is to recover spatial details lost during the downsampling in the encoder, facilitating precise localization and detection of salient objects. \\
\subsubsection{\textbf{Attention Gate Module}} \label{AG}
\normalsize Attention Gates (AGs) have demonstrated remarkable effectiveness in capturing crucial regions, diminishing feature responses in irrelevant background areas, and eliminating the need to crop a region of interest (ROI) within an image. This is particularly important for the task of visual saliency detection. The integration of AGs into the conventional U-Net architecture enhances the model's ability to emphasize salient features transmitted through skip connections.
Given a skip connection feature \(F_{s} \in R^{C \times H \times W}\), where C is the number of channels, H and W are the height and width of F, we first apply a convolution layer, batch norm, and ReLu activation function to obtain a key feature \(K\), and let Q be the input from the previous layer or the gating signal obtained by applying a convolution layer followed by a batch normalization and relu activation to the input gate feature \(F_{g}\). The attention coefficient \( \alpha \) is obtained by applying a relu function to the element-wise sum of \(Q \) and \(K\). The final attention coefficient value \(V\) is obtained by feeding a convolution layer, batch normalization, and sigmoid activation function to the attention coefficient \(\alpha\). 
\begin{figure*}[H]
    \centering
    \includegraphics[width=0.99\textwidth]{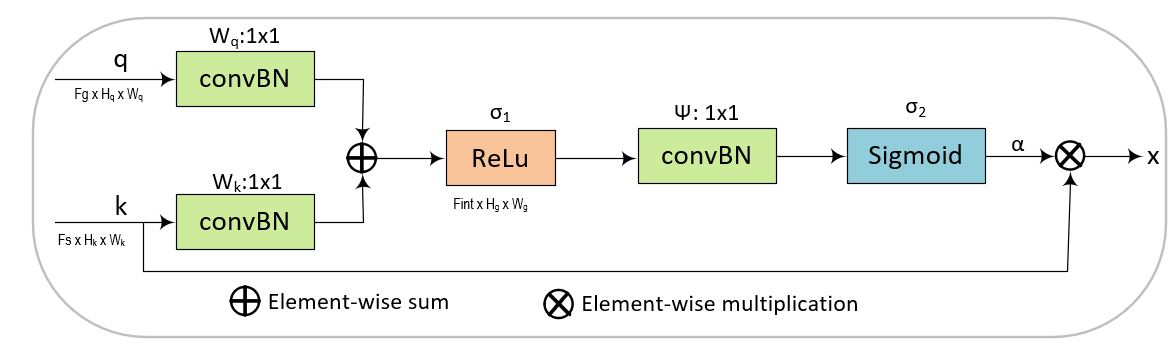}
    \caption{ Diagram of the additive attention gate (AG) module.}
    \label{fig:attention gate}
\end{figure*}
Finally, the attention coefficient value V and the skip connection feature map are multiplied element-wise to produce the final attention gate output $\hat{x}^{l}_{i, c}$, which is calculated as equation \ref{agoutput}
\begin{align}
q^{l}_{att} &= \Psi(\sigma_{1}(W^{T}_{q} \times q^{l}_{i} + W^{T}_{k} \times K_{i} + b_{k})) + b_{\Psi} \\
\alpha_{i}^{l} &= \sigma_{2}(q^{l}_{att}(q^{l}_{i}, K_{i}; \Theta_{att})) \\
\hat{x}_{i, c}^l &= \alpha_{i}^{l} \cdot K_{i, c} \label{agoutput}
\end{align}
where \( \sigma_{2}(x_{i}, c) = \frac{1}{1 + exp^{(-x_{i}, c)}} \) represents the sigmoid activation function. 
Thus, AG is defined by a set of parameter set $\theta_{att}$, which includes linear transformations $ W_{k} \in R^{F_{l} \times F_{int}}, W_{q} \in R^{F_{g} \times F_{int}}, \Psi \in R^{F_{int} \times 1}$ and bias terms $b_{\Psi} \in R, b_{k} \in R^{F_{int}}$. The linear transformation can be computed using channel-wise $1 \times 1 $ convolutions for the input tensors. \\

\subsubsection{\textbf{Saliency Fusion Module}}\label{sfm} 
\normalsize The saliency map fusion module serves as a pivotal component in generating saliency probability maps. Similar to the methodology in ~\cite{qin2020u2}, our model undertakes a multi-stage approach. Initially, it generates four side output saliency probability maps, denoted as \( S(1)_{\text{side}}, \\
S(2)_{\text{side}}, S(3)_{\text{side}}, \) and \( S(4)_{\text{side}} \), originating from the respective stages decoder1, decoder2, decoder3, and decoder4. This generation is facilitated by a \(3 \times 3\) convolution layer, followed by a sigmoid activation function. Subsequently, the convolution outputs prior to sigmoid functions of these side output saliency maps are upsampled to have the same size as the input image. The integration of these saliency maps is accomplished through a concatenation operation, followed by a \(1 \times 1\) convolution layer and a sigmoid function. The result of this fusion process is the final saliency map \( S_{\text{fuse}} \) (depicted in the bottom right of Figure ~\ref{fig:model_arc}).

Mathematically, the saliency probability maps at each stage are generated as follows:
\begin{align}
S(i)_{\text{side}} = \sigma(\text{Conv}(i)(X))
\end{align}
where \( i \) represents the stage (1, 2, 3, or 4), \( \sigma \) denotes the sigmoid function, \( \text{Conv}(i) \) is the convolution operation at stage \( i \), and \( X \) is the decoder's output feature map.
The side outputs are then upsampled and concatenated to generate the final saliency map \( S_{\text{fuse}} \):
\begin{align}
S_{\text{fuse}} = \sigma(\text{Conv}_{\text{fuse}}(\text{Concat}(S(i)_{\text{side}})))
\end{align}
where \text{Concat} represents the concatenation operation, \( \text{Conv}_{\text{fuse}} \) is the \(1 \times 1\) convolution layer specific to the fusion process, and \( \sigma \) represents the sigmoid function.

\subsection{\textbf{Network Supervision}}
\label{lossfn}
\normalsize Loss functions play a significantly important role in optimizing a saliency detection model. One of the most widely employed loss functions for binary classification problems is the binary cross-entropy (BCE) loss \cite{de2005tutorial}. For visual saliency detection, it measures the dissimilarity between the predicted saliency map and the ground truth in a binary classification setting. 

We use a deep supervision approach similar to that in \cite{qin2020u2}, which has demonstrated efficacy. Our training loss is formulated as follows:

\begin{align}
\mathcal{L} = \sum^{M}_{m=1}w^{m}_{side}l^{m}_{side} + w_{fuse}l_{fuse} \label{eq:total_loss}
\end{align}
The total loss comprises two components. The first component is the loss associated with the side output saliency maps, denoted as $l^{m}_{side}$, where $m$ represents the four supervision stages (Sup1, Sup2, Sup3, and Sup4) shown in Figure \ref{fig:model_arc}. The second component is the loss of the final fusion output saliency map, represented by $l_{fuse}$. The weights assigned to these loss terms are $w^{m}_{side}$ and $w_{fuse}$, respectively.\\
We compute the loss for each term $\mathit{l}$ using the conventional binary cross-entropy to calculate the pixel level comparison between the predicted saliency map and the ground truth.

\begin{align}
\resizebox{0.45\textwidth}{!}{$
\mathit{l} = - \sum^{(H, W)}_{(x, y)}[G_{(x, y)} \log P_{(x, y)} + (1- P_{(x, y)})\log (1-P_{(x, y)})]$}
\end{align}

where (H, W) is the height and width of the image, and (x, y) is the coordinate of a pixel. The ground truth and predicted saliency probability map's pixel values are represented by the symbols G(x, y) and P(x, y), respectively. The goal of the training procedure is to reduce the total loss $\mathcal{L}$ of \eqref{eq:total_loss}. We select the fusion output $l_{fuse}$ as our final saliency map during the testing process. 

\section{Experimental Results} \label{Experimental results}
\normalsize 
\begin{table*}[h]
  \caption{Comparison of the proposed method and 4 other methods on DUT-OMRON, DUTS-TE, and ECSSD datasets, using $MAE(\downarrow)$, F-measure $F_{\beta}(\uparrow)$, structure measure $S_{m}(\uparrow)$, and  e-measure $E_{m}(\uparrow)$ as evaluation metrics. The best values are highlighted in bold. }
 \small
 \resizebox{\textwidth}{!}
  {
   \begin{tabularx}{1.1\textwidth}{c *{3}{X} *{3}{X} *{3}{X} *{3}{X}}
    \toprule
    Dataset & \multicolumn{4}{c}{\hspace{1pt}DUT-OMRON} & \multicolumn{4}{c}{\hspace{1pt}DUTS-TE} & \multicolumn{4}{c}{\hspace{1pt}ECSSD} \\
    \midrule
    metrics & MAE & $F_{\beta}$ & $ S_{m} $ & $E_{m}$ & MAE & $F_{\beta}$ & $ S_{m} $ & $E_{m}$ & MAE & $F_{\beta}$ & $ S_{m} $ & $E_{m}$ \\
    \midrule
    DCL   & 0.132 & 0.734 & 0.758 & 0.763 & 0.174 & 0.771 & 0.776 & 0.764 & 0.078 & 0.910 & 0.883 & 0.888 \\
    RFCN  & 0.110 & 0.742 & 0.764 & 0.778 & 0.09 & 0.784 & 0.794 & 0.839 & 0.107 & 0.890 & 0.852 & 0.876 \\
    UCF   & 0.132 & 0.734 & 0.758 & 0.763 & 0.117 & 0.771 & 0.776 & 0.764 & 0.078 & 0.910 & 0.883 & 0.888 \\
    Amulet & 0.098 & \textbf{0.743} & \textbf{0.780} & 0.784 & 0.085 & 0.778 & 0.802 & 0.797 & \textbf{0.059} & \textbf{0.912} &\textbf{0.893} & \textbf{0.911} \\
    ours  & \textbf{0.080} & 0.722 & 0.755 & \textbf{0.811} & \textbf{0.061} & {\textbf{0.794}} & \textbf{0.809} & \textbf{0.856} & 0.063 & 0.895 & 0.862 & 0.892 \\
    \bottomrule
    \end{tabularx}%
    }
  \label{tab:table1}%
\end{table*}%

\begin{table*}[h]
  \centering
  \caption{Comparison of the proposed method and 4 other methods on HKU-IS, PASCAL-S, and SOD datasets, using  $MAE(\downarrow)$, F-measure $F_{\beta}(\uparrow)$, structure measure $S_{m}(\uparrow)$, and  e-measure $E_{m}(\uparrow)$ as evaluation metrics. The best values are highlighted in bold.}
  \small
  \resizebox{\textwidth}{!}
  {
  \begin{tabularx}{1.1\textwidth}{c *{3}{X} *{3}{X} *{3}{X} *{3}{X}}
    \toprule
    Dataset & \multicolumn{4}{c}{\hspace{1pt}HKU-IS} & \multicolumn{4}{c}{\hspace{1pt}PASCAL-S} & \multicolumn{4}{c}{\hspace{1pt}SOD} \\
    \midrule
    metrics & MAE & $F_{\beta}$ & $ S_{m} $ & $E_{m}$ & MAE & $F_{\beta}$ & $ S_{m} $ & $E_{m}$ & MAE & $F_{\beta}$ & $ S_{m} $ & $E_{m}$ \\
    \midrule
    DCL   & 0.074 & 0.886 & 0.866 & 0.891 & 0.126 & 0.824 & 0.803 & 0.785 & 0.164 & 0.798 & 0.754 & 0.755 \\
    RFCN  & 0.089 & 0.893 & 0.859 & 0.906 & 0.132 & 0.824 & 0.798 & 0.807 & 0.169 & 0.797 & 0.732 & 0.778 \\
    UCF   & 0.074 & 0.886 & 0.866 & 0.891 & 0.126 & 0.824 & 0.803 & 0.785 & 0.164 & 0.798 & 0.754 & 0.755 \\
    Amulet & 0.052 & 0.895 & 0.882 & 0.912 & 0.098 & \textbf{0.833} & \textbf{0.819} & 0.827 & 0.0141 & \textbf{0.802} & \textbf{0.759} & \textbf{0.791} \\
    Ours  & \textbf{0.044} &\textbf{0.896} &\textbf{0.885}& \textbf{0.927} & \textbf{0.091} & 0.814 & 0.801 & \textbf{0.834} & \textbf{0.137} & 0.798 & 0.718 & 0.759 \\
    \bottomrule
  \end{tabularx}
  }
  \label{tab:table2}
\end{table*}
\subsection{\textbf{Datasets}} \label{dataset}

\normalsize \textbf{Training dataset:} We train our model using DUTS-TR dataset,which is a subset of the DUTS dataset \cite{wang2017}. DUTS-TR is curated from the training and validation sets of ImageNet DET \cite{deng2009imagenet}, and comprises a total of 10,553 images, each with its corresponding ground truth. DUTS is the largest and most widely used dataset for saliency detection. We performed horizontally flipping data augmentation technique, resulting in 21,106 images for training.  \\
\textbf{Evaluation dataset:}
We use the following six widely used saliency detection datasets inorder to evaluate the detection performances of our model.\\
\textbf{ECSSD \cite{7182346}}: The ECSSD (Extended Complex Scene Saliency Dataset) contains semantically significant yet structurally complex and challenging images. This dataset contains 1000 natural images with carefully annotated ground truth saliency masks.\\
\textbf{PASCAL-S \cite{li2014secrets}:} This dataset was collected on 8 subjects with a 3-second viewing time and the utilization of the Eyelink eye tracker collected from the PASCAL VOC (Visual Object Classes 2010) \cite{Everingham2010ThePV} validation dataset. This dataset contains 850 images featuring multiple salient objects within their scenes, providing a rich and diverse visual context.   \\
\textbf{HKU-IS \cite{7299184}:} The HKU-IS dataset is a more challenging benchmark for visual saliency detection, aimed at advancing the research and evaluating the performance of visual saliency models. This dataset comprises 4447 challenging images, featuring high-quality pixel-wise annotations with characteristics of either low contrast or presence of multiple salient objects.\\
\textbf{DUT-OMRON \cite{yang2013saliency}:} DUT-OMRON comprises 5,168 high-quality nature images meticulously chosen from more than 140,000 images. These images possess dimensions of either 400*x or x*400 pixel dimensions, where x is less than 400. Notably, each image features one or more salient objects set against a relatively complex background. \\
\textbf{DUTS-TE:} DUTS-TE is the test sets of DUTS dataset, which comprises 5, 019 test images sourced from the ImageNet DET test set and the SUN dataset \cite{xiao2010sun}. This dataset contains highly challenging scenarios for the evaluation of saliency detection models. \\
\textbf{SOD \cite{movahedi2010design}:} SOD comprises salient object boundaries derived from the Berkeley Segmentation Dataset (BSD)        \cite{martin2001database}. It consists of 300 particularly challenging images, initially intended for image segmentation. \\
 
\subsection{\textbf{Evaluation Metrics}} \label{eval metric}
\normalsize The probability maps that are produced by deep salient object algorithms often have the same dimension as the input images. In predicted saliency maps, each pixel has a value between 0 and 1 (or [0, 255]). The ground truths are often binary masks, where each pixel is either 0 or 1 (or 0 and 255), with 1 denoting the pixels of the foreground salient object and 0 denoting the background.

To comprehensively evaluate the performance of our model and the quality of the predicted saliency maps against the actual saliency masks, we used the following four evaluating measures: (1) Mean Absolute Error (MAE) \cite{6247743},(2) maximal F-measure \((maxF \beta)\) \cite{achanta2009frequency},  (3) structure measure (Sm)  \cite{fan2017structure}, and (4) enhanced alignment measure (Em) \cite{fan2018enhanced}. The detailed descriptions of these measures are presented below. \\ 
\textbf{1. F-measure} \\
F-measure comprehensively evaluates both precision and recall  as:
\begin{align}
F_{\beta} = \frac{(1 + \beta_{2})Precision \times Recall}{\beta_{2}Precision + Recall} \label{eq:f-m}
\end{align}
Since the rate of recall is not as important as precision, $ \beta_{2} $ is empirically set to 0.3 to emphasize precision more. \\
\textbf{2. Mean Absolute Error(MAE})\\
MAE, or Mean Absolute Error, represents the average difference per pixel between a predicted saliency map and its corresponding ground truth mask. It is used as a metric to accurately assess false negative pixels. 
\begin{align}
MAE = \frac{1}{H \times W} \sum^{H}_{x=1} \sum^{W}_{y=1}|P(x, y)- G(x, y)| \label{eq:mae}
\end{align}
where P and G are the probability map of saliency detection and the corresponding ground truth, respectively, and (H, W) and (x, y) are the (height, width) and the pixel coordinates.
A lower MAE value signifies a high degree of similarity between the ground truth and the predicted saliency map.\\
\textbf{3. Structure Measure (sm)}\\
S-measure (Sm) assesses the structural similarity between the predicted non-binary saliency map and the binary ground truth. It is defined as the weighted sum of region-aware Sr and object-aware So structural similarity:
\begin{align}
S = (1 - \alpha)Sr + \alpha So  \label{eq:sm}
\end{align}
Typically, $ \alpha $ is set to 0.5.\\
\textbf{4. Enhanced alignment measure (em)}\\
Enhanced alignment measure (em) incorporates both local pixel matching information and image-level statics by combining local pixel values and the image-level mean or global average value in a single term.
\begin{align}
Q_{FM} =  \frac{1}{h \times w} \sum^{h}_{x=1}\sum^{w}_{y=1} \phi_{FM}(x, y)\label{eq:em}
\end{align}
Where h and w are the height and width of the saliency map, respectively. $ \phi_{FM} $ is enhanced alignment matrix, reflecting the correlation between P and G after subtracting their global means, respectively.

\subsection{\textbf{Implementation Details}} \label{Implementation}

\normalsize The proposed network is implemented using the PyTorch framework, and training and testing are performed on a NVIDIA GeForce RTX 4070Ti GPU with 12 GB of video memory. The training dataset consists of 10,553 images from the DUTS-TR subset of DUTS \cite{wang2017}. To augment the dataset, each image is horizontally flipped, resulting in a doubled training set with 21,106 images. Prior to feeding the images into the network, they are resized to 320 x 320 and then cropped to 288 x 288 during training. Model optimization employs the Adam optimizer with default hyperparameter values $ (lr=1e-3, betas=(0.9, 0.999), eps=1e-8, weight\_decay=0)$. The network is trained for approximately 500,000 iterations with a batch size of 12 to ensure convergence of the loss. While testing, the input images are first resized to 320 x 320 before being inputted into the trained network. The resulting predicted saliency map is then restored to its original dimensions through bilinear interpolation.
\begin{figure}[ht]
    \centering
    \includegraphics[width=0.5\textwidth]{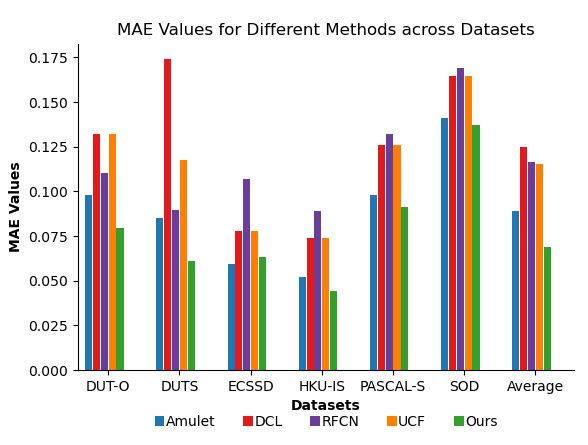}
    \caption{ MAE values of Amulet, DCL, RFCN, UCF, and the proposed SalFAU-Net model across six evaluation datasets and the average MAE values of each method across all datasets. A lower MAE value signifies superior performance.}
    \label{fig:avgmae_graph}
\end{figure}


\subsection{Comparison with Other Methods} \label{comparison with other methods}
\normalsize In this section, we evaluate the effectiveness of the proposed model through both qualitative and quantitative analysis. We perform experiments to compare its performance with that of other models, utilizing four evaluation metrics, namely, MAE, F-measure, s-measure, and e-measure.
We compare the results of the proposed method with some FCN-based methods, including Amulet \cite{zhang2017amulet}, DCL \cite{li2016deep}, RFCN \cite{wang2016saliency}, and UCF \cite{zhang2017learning}.
\subsubsection{Quantitative Comparison} \label{quant comp}
The quantitative results on the six evaluation datasets using the four evaluation metrics are reported in Table \ref{tab:table1} and Table \ref{tab:table2}. Based on the results, it is evident that our proposed method outperforms the benchmark methods on DUTS-TE and HKU-IS datasets in all evaluation metrics. On the DUT-OMRON dataset, our model achieves impressive results, outperforming other methods with the best MAE values. Furthermore, on the ECSSD dataset, our model demonstrates competitive performance with the second lowest MAE value of 0.063, surpassed only by Amulet with a slightly lower MAE of 0.059. Furthermore, we calculate the average MAE values for each method across the six datasets. Impressively, our proposed method achieves the lowest average MAE value of 0.068, indicating superior performance compared to other methods. Figure \ref{fig:avgmae_graph} presents the results of the average MAE values of each method, which clearly demonstrate that our model outperforms the comparison algorithms based on the average MAE value.
\begin{figure*}[ht]
    \centering
    \includegraphics[width=\textwidth, height=0.5\textheight]{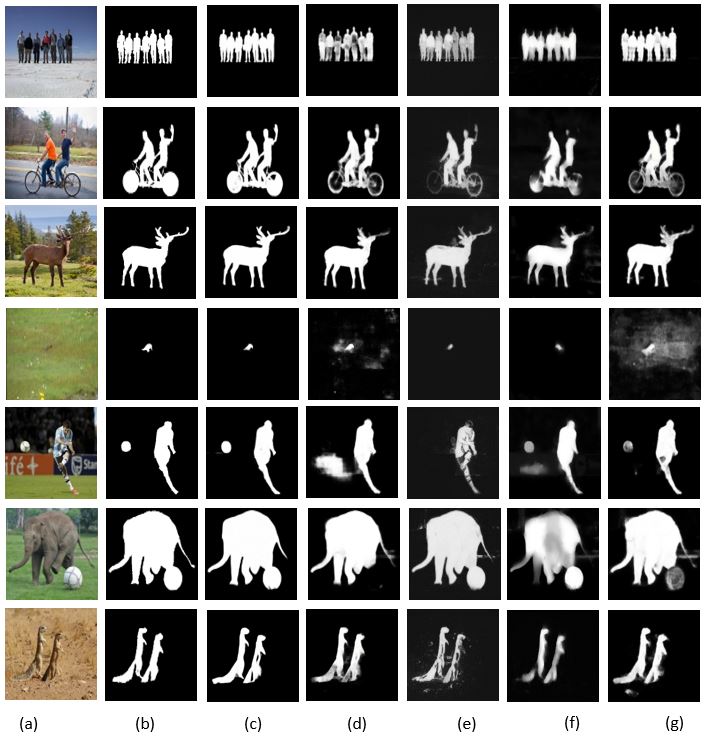}
    \caption{Qualitative comparison of the proposed method with four other SOTA methods: (a) original image, (b) GT, (c) Ours, (d)
Amulet, (e) DCL, (f) RFCN, (g) UCF}
    \label{fig:Salmap}
\end{figure*}
\subsubsection{Qualitative Comparison} \label{qual result}
\normalsize In addition to quantitative evaluations, we present predicted saliency maps generated by the proposed method and the comparison methods in Figure \ref{fig:Salmap}. The images in the first and second columns of Figure \ref{fig:Salmap} represent the original input images and their corresponding ground truth saliency maps, respectively. The third column showcases the predicted saliency maps of our proposed method, while the fourth, fifth, sixth, and seventh columns exhibit the results of the comparison methods. The first two rows depict scenarios with multiple salient objects; the third row showcases a single large salient object; the fourth row contains small objects; and the fifth and sixth rows depict images with both small and large salient objects. The last row features relatively low-contrast salient objects.  As we can see from Figure \ref{fig:Salmap}, the results demonstrate that SalFAU-Net generates saliency maps more accurately for different challenging scenes, while the comparison methods generate incomplete or noisy saliency maps.

From the qualitative and quantitative results presented above, it is evident that our proposed method yields competitive results in tackling the challenge of visual saliency detection. These findings also highlight the crucial role of attention mechanisms in enhancing the effectiveness of the visual saliency models, as the proposed model places significant emphasis on extracting highly representative features while effectively eliminating unwanted or noisy features. This emphasis on attention mechanisms not only contributes to the competitiveness of our approach but also enhances the overall performance by prioritizing the extraction of relevant or salient visual information. 
\subsection{\textbf{Failure cases}} \label{failure cases}
\begin{figure}[H]
    \includegraphics[width=0.5\textwidth]{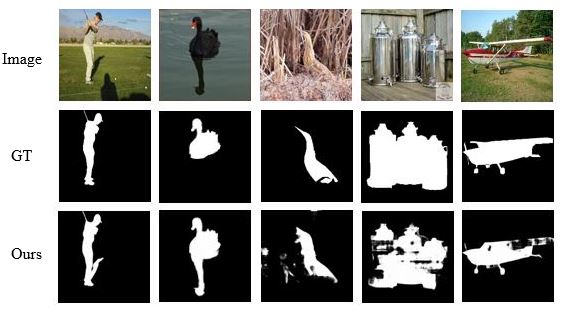}
    \caption{ Failure cases of the proposed method.}
    \label{fig:Failure cases}
\end{figure}

\normalsize The proposed method demonstrated effective salient object detection in most cases. However, there are some cases where it exhibits limitations. Figure \ref{fig:Failure cases} shows some failure cases for the proposed method. In the first column of Figure \ref{fig:Failure cases}, the presence of the shadow of the person is erroneously detected as a salient object. This is because the presence of shadows in salient objects may cause a decrease in the visibility or distinguishability of salient objects, making them harder to detect accurately. In the second column, the reflection of the duck is identified as a salient object, which is caused by reflections, which can create distracting image regions, potentially diverting attention away from the true salient objects. The third and fourth columns depict situations where the salient objects have low contrast, causing difficulty for our model in accurate saliency detection. In the last column, although most part of the airplane is detected, the model fails to capture its entirety.

In general, these images are very challenging for most deep learning models to detect accurately. These challenges arise due to the sensitivity of deep learning models to factors such as shadows, reflections, and low contrasts in salient objects. In the future, we will carry out further research aiming to address these problems and develop more accurate saliency detection models. 
\newpage
\section{Conclusion}\label{conc}
\normalsize In this paper, we proposed Saliency Fusion Attention U-Net (SalFAU-Net) as an approach for visual saliency detection tasks. Our method integrates a saliency fusion module into each decoder block of the Attention U-Net model, which enables efficient generation of saliency maps. 
The use of an attention gate module in our method facilitates selective focus on informative regions and suppression of non-salient regions within an image. The comprehensive evaluation across six diverse SOD datasets, both quantitatively and qualitatively, underscores the effectiveness of our proposed method compared to the benchmark methods. SalFAU-Net not only showcases competitive performance but also highlights the potential of attention-based models in advancing the capabilities of saliency detection models.

\bibliography{salfaunet_references}

\begin{thebibliography}{10}

\bibitem{achanta2008salient}
Radhakrishna Achanta, Francisco Estrada, Patricia Wils, and Sabine S{\"u}sstrunk, `Salient region detection and segmentation', in {\em Computer Vision Systems: 6th International Conference, ICVS 2008 Santorini, Greece, May 12-15, 2008 Proceedings 6}, pp. 66--75. Springer, (2008).

\bibitem{achanta2009frequency}
Radhakrishna Achanta, Sheila Hemami, Francisco Estrada, and Sabine Susstrunk, `Frequency-tuned salient region detection', in {\em 2009 IEEE conference on computer vision and pattern recognition}, pp. 1597--1604. IEEE, (2009).

\bibitem{borji2014salient}
Ali Borji, `What is a salient object? a dataset and a baseline model for salient object detection', {\em IEEE Transactions on Image Processing}, {\bf 24}(2),  742--756, (2014).

\bibitem{borji2011scene}
Ali Borji and Laurent Itti, `Scene classification with a sparse set of salient regions', in {\em 2011 IEEE International Conference on Robotics and Automation}, pp. 1902--1908. IEEE, (2011).

\bibitem{de2005tutorial}
Pieter-Tjerk De~Boer, Dirk~P Kroese, Shie Mannor, and Reuven~Y Rubinstein, `A tutorial on the cross-entropy method', {\em Annals of operations research}, {\bf 134},  19--67, (2005).

\bibitem{deng2009imagenet}
Jia Deng, Wei Dong, Richard Socher, Li-Jia Li, Kai Li, and Li~Fei-Fei, `Imagenet: A large-scale hierarchical image database', in {\em 2009 IEEE conference on computer vision and pattern recognition}, pp. 248--255. Ieee, (2009).

\bibitem{donoser2009saliency}
Michael Donoser, Martin Urschler, Martin Hirzer, and Horst Bischof, `Saliency driven total variation segmentation', in {\em 2009 IEEE 12th International Conference on Computer Vision}, pp. 817--824. IEEE, (2009).

\bibitem{duan2022saliency}
Huiyu Duan, Wei Shen, Xiongkuo Min, Danyang Tu, Jing Li, and Guangtao Zhai, `Saliency in augmented reality', in {\em Proceedings of the 30th ACM International Conference on Multimedia}, pp. 6549--6558, (2022).

\bibitem{Everingham2010ThePV}
Mark Everingham, Luc~Van Gool, Christopher K.~I. Williams, John~M. Winn, and Andrew Zisserman, `The pascal visual object classes (voc) challenge', {\em International Journal of Computer Vision}, {\bf 88},  303--338, (2010).

\bibitem{fan2017structure}
Deng-Ping Fan, Ming-Ming Cheng, Yun Liu, Tao Li, and Ali Borji, `Structure-measure: A new way to evaluate foreground maps', in {\em Proceedings of the IEEE international conference on computer vision}, pp. 4548--4557, (2017).

\bibitem{fan2018enhanced}
Deng-Ping Fan, Cheng Gong, Yang Cao, Bo~Ren, Ming-Ming Cheng, and Ali Borji, `Enhanced-alignment measure for binary foreground map evaluation', {\em arXiv preprint arXiv:1805.10421}, (2018).

\bibitem{fu2017object}
Huazhu Fu, Dong Xu, and Stephen Lin, `Object-based multiple foreground segmentation in rgbd video', {\em IEEE Transactions on Image Processing}, {\bf 26}(3),  1418--1427, (2017).

\bibitem{gong2022improved}
Xiaoran Gong, Letu Qingge, Qing Liu, and Pei Yang, `Improved u-net-like network for visual saliency detection based on pyramid feature attention', {\em Wireless Communications and Mobile Computing}, {\bf 2022}, (2022).

\bibitem{han2019convolutional}
Le~Han, Xuelong Li, and Yongsheng Dong, `Convolutional edge constraint-based u-net for salient object detection', {\em IEEE Access}, {\bf 7},  48890--48900, (2019).

\bibitem{hu2020sac}
Xiaowei Hu, Chi-Wing Fu, Lei Zhu, Tianyu Wang, and Pheng-Ann Heng, `Sac-net: Spatial attenuation context for salient object detection', {\em IEEE Transactions on Circuits and Systems for Video Technology}, {\bf 31}(3),  1079--1090, (2020).

\bibitem{ji2021cnn}
Yuzhu Ji, Haijun Zhang, Zhao Zhang, and Ming Liu, `Cnn-based encoder-decoder networks for salient object detection: A comprehensive review and recent advances', {\em Information Sciences}, {\bf 546},  835--857, (2021).

\bibitem{lei2019dilated}
Xinyu Lei, Hongguang Pan, and Xiangdong Huang, `A dilated cnn model for image classification', {\em IEEE Access}, {\bf 7},  124087--124095, (2019).

\bibitem{7299184}
Guanbin Li and Y.~Yu, `Visual saliency based on multiscale deep features', in {\em 2015 IEEE Conference on Computer Vision and Pattern Recognition (CVPR)}, pp. 5455--5463, Los Alamitos, CA, USA, (jun 2015). IEEE Computer Society.

\bibitem{li2016deep}
Guanbin Li and Yizhou Yu, `Deep contrast learning for salient object detection', in {\em Proceedings of the IEEE Conference on Computer Vision and Pattern Recognition}, pp. 478--487, (2016).

\bibitem{li2015inner}
Hongyang Li, Huchuan Lu, Zhe Lin, Xiaohui Shen, and Brian Price, `Inner and inter label propagation: salient object detection in the wild', {\em IEEE Transactions on Image Processing}, {\bf 24}(10),  3176--3186, (2015).

\bibitem{li2020stacked}
Junxia Li, Zefeng Pan, Qingshan Liu, and Ziyang Wang, `Stacked u-shape network with channel-wise attention for salient object detection', {\em IEEE Transactions on Multimedia}, {\bf 23},  1397--1409, (2020).

\bibitem{li2013saliency}
Xiaohui Li, Huchuan Lu, Lihe Zhang, Xiang Ruan, and Ming-Hsuan Yang, `Saliency detection via dense and sparse reconstruction', in {\em Proceedings of the IEEE international conference on computer vision}, pp. 2976--2983, (2013).

\bibitem{li2014secrets}
Yin Li, Xiaodi Hou, Christof Koch, James~M Rehg, and Alan~L Yuille, `The secrets of salient object segmentation', in {\em Proceedings of the IEEE conference on computer vision and pattern recognition}, pp. 280--287, (2014).

\bibitem{liu2021review}
Xiangbin Liu, Liping Song, Shuai Liu, and Yudong Zhang, `A review of deep-learning-based medical image segmentation methods', {\em Sustainability}, {\bf 13}(3),  1224, (2021).

\bibitem{long2015fully}
Jonathan Long, Evan Shelhamer, and Trevor Darrell, `Fully convolutional networks for semantic segmentation', in {\em Proceedings of the IEEE conference on computer vision and pattern recognition}, pp. 3431--3440, (2015).

\bibitem{martin2001database}
David Martin, Charless Fowlkes, Doron Tal, and Jitendra Malik, `A database of human segmented natural images and its application to evaluating segmentation algorithms and measuring ecological statistics', in {\em Proceedings Eighth IEEE International Conference on Computer Vision. ICCV 2001}, volume~2, pp. 416--423. IEEE, (2001).

\bibitem{movahedi2010design}
Vida Movahedi and James~H Elder, `Design and perceptual validation of performance measures for salient object segmentation', in {\em 2010 IEEE computer society conference on computer vision and pattern recognition-workshops}, pp. 49--56. IEEE, (2010).

\bibitem{oktay2018attention}
Ozan Oktay, Jo~Schlemper, Loic~Le Folgoc, Matthew Lee, Mattias Heinrich, Kazunari Misawa, Kensaku Mori, Steven McDonagh, Nils~Y Hammerla, Bernhard Kainz, et~al., `Attention u-net: Learning where to look for the pancreas', {\em arXiv preprint arXiv:1804.03999}, (2018).

\bibitem{pal2020looking}
Anwesan Pal, Sayan Mondal, and Henrik~I Christensen, `" looking at the right stuff"-guided semantic-gaze for autonomous driving', in {\em Proceedings of the IEEE/CVF Conference on Computer Vision and Pattern Recognition}, pp. 11883--11892, (2020).

\bibitem{6247743}
Federico Perazzi, Philipp Krähenbühl, Yael Pritch, and Alexander Hornung, `Saliency filters: Contrast based filtering for salient region detection', in {\em 2012 IEEE Conference on Computer Vision and Pattern Recognition}, pp. 733--740, (2012).

\bibitem{qin2014integration}
Chanchan Qin, Guoping Zhang, Yicong Zhou, Wenbing Tao, and Zhiguo Cao, `Integration of the saliency-based seed extraction and random walks for image segmentation', {\em Neurocomputing}, {\bf 129},  378--391, (2014).

\bibitem{qin2020u2}
Xuebin Qin, Zichen Zhang, Chenyang Huang, Masood Dehghan, Osmar~R Zaiane, and Martin Jagersand, `U2-net: Going deeper with nested u-structure for salient object detection', {\em Pattern recognition}, {\bf 106},  107404, (2020).

\bibitem{ren2013region}
Zhixiang Ren, Shenghua Gao, Liang-Tien Chia, and Ivor Wai-Hung Tsang, `Region-based saliency detection and its application in object recognition', {\em IEEE Transactions on Circuits and Systems for Video Technology}, {\bf 24}(5),  769--779, (2013).

\bibitem{ronneberger2015u}
Olaf Ronneberger, Philipp Fischer, and Thomas Brox, `U-net: Convolutional networks for biomedical image segmentation', in {\em Medical Image Computing and Computer-Assisted Intervention--MICCAI 2015: 18th International Conference, Munich, Germany, October 5-9, 2015, Proceedings, Part III 18}, pp. 234--241. Springer, (2015).

\bibitem{rutishauser2004bottom}
Ueli Rutishauser, Dirk Walther, Christof Koch, and Pietro Perona, `Is bottom-up attention useful for object recognition?', in {\em Proceedings of the 2004 IEEE Computer Society Conference on Computer Vision and Pattern Recognition, 2004. CVPR 2004.}, volume~2, pp. II--II. IEEE, (2004).

\bibitem{7182346}
Jianping Shi, Qiong Yan, Li~Xu, and Jiaya Jia, `Hierarchical image saliency detection on extended cssd', {\em IEEE Transactions on Pattern Analysis and Machine Intelligence}, {\bf 38}(4),  717--729, (2016).

\bibitem{ullah2020brief}
Inam Ullah, Muwei Jian, Sumaira Hussain, Jie Guo, Hui Yu, Xing Wang, and Yilong Yin, `A brief survey of visual saliency detection', {\em Multimedia Tools and Applications}, {\bf 79},  34605--34645, (2020).

\bibitem{wang2017}
Lijun Wang, Huchuan Lu, Yifan Wang, Mengyang Feng, Dong Wang, Baocai Yin, and Xiang Ruan, `Learning to detect salient objects with image-level supervision', in {\em CVPR}, (2017).

\bibitem{wang2016saliency}
Linzhao Wang, Lijun Wang, Huchuan Lu, Pingping Zhang, and Xiang Ruan, `Saliency detection with recurrent fully convolutional networks', in {\em European Conference on Computer Vision}, pp. 825--841. Springer, (2016).

\bibitem{xiao2010sun}
Jianxiong Xiao, James Hays, Krista~A Ehinger, Aude Oliva, and Antonio Torralba, `Sun database: Large-scale scene recognition from abbey to zoo', in {\em 2010 IEEE computer society conference on computer vision and pattern recognition}, pp. 3485--3492. IEEE, (2010).

\bibitem{xie2021oriented}
Xingxing Xie, Gong Cheng, Jiabao Wang, Xiwen Yao, and Junwei Han, `Oriented r-cnn for object detection', in {\em Proceedings of the IEEE/CVF international conference on computer vision}, pp. 3520--3529, (2021).

\bibitem{xie2012bayesian}
Yulin Xie, Huchuan Lu, and Ming-Hsuan Yang, `Bayesian saliency via low and mid level cues', {\em IEEE Transactions on Image Processing}, {\bf 22}(5),  1689--1698, (2012).

\bibitem{xu2015show}
Kelvin Xu, Jimmy Ba, Ryan Kiros, Kyunghyun Cho, Aaron Courville, Ruslan Salakhudinov, Rich Zemel, and Yoshua Bengio, `Show, attend and tell: Neural image caption generation with visual attention', in {\em International conference on machine learning}, pp. 2048--2057. PMLR, (2015).

\bibitem{yan2013hierarchical}
Qiong Yan, Li~Xu, Jianping Shi, and Jiaya Jia, `Hierarchical saliency detection', in {\em Proceedings of the IEEE conference on computer vision and pattern recognition}, pp. 1155--1162, (2013).

\bibitem{yang2013saliency}
Chuan Yang, Lihe Zhang, Huchuan Lu, Xiang Ruan, and Ming-Hsuan Yang, `Saliency detection via graph-based manifold ranking', in {\em Proceedings of the IEEE conference on computer vision and pattern recognition}, pp. 3166--3173, (2013).

\bibitem{zhang2017amulet}
Pingping Zhang, Dong Wang, Huchuan Lu, Hongyu Wang, and Xiang Ruan, `Amulet: Aggregating multi-level convolutional features for salient object detection', in {\em Proceedings of the IEEE International Conference on Computer Vision}, pp. 202--211, (2017).

\bibitem{zhang2017learning}
Pingping Zhang, Dong Wang, Huchuan Lu, Hongyu Wang, and Baocai Yin, `Learning uncertain convolutional features for accurate saliency detection', in {\em Computer Vision (ICCV), 2017 IEEE International Conference on}, pp. 212--221. IEEE, (2017).

\bibitem{zhang2020attention}
Qing Zhang, Yanjiao Shi, and Xueqin Zhang, `Attention and boundary guided salient object detection', {\em Pattern Recognition}, {\bf 107},  107484, (2020).

\bibitem{zhao2019pyramid}
Ting Zhao and Xiangqian Wu, `Pyramid feature attention network for saliency detection', in {\em Proceedings of the IEEE/CVF conference on computer vision and pattern recognition}, pp. 3085--3094, (2019).

\end{thebibliography}
\end{document}